\documentclass[10pt,twocolumn,letterpaper]{article}

\usepackage[pagenumbers]{wacv} 

\usepackage{graphicx}
\usepackage{amsmath}
\usepackage{amssymb}
\usepackage{booktabs}

\usepackage{booktabs}       
\usepackage{amsfonts}       
\usepackage{nicefrac}       
\usepackage{microtype}      
\usepackage{xcolor}         
\usepackage{graphicx}
\usepackage{graphbox}
\usepackage{wrapfig}
\usepackage{amsmath}
\usepackage[font=small]{caption}
\usepackage[font=footnotesize]{subcaption}
\usepackage{floatrow}
\usepackage{pifont}
\usepackage{makecell}
\usepackage{multicol,multirow}
\usepackage{color, colortbl}
\definecolor{Gray}{gray}{0.9}
\usepackage[export]{adjustbox}

\definecolor{Gray}{gray}{0.9}

\usepackage[pagebackref,breaklinks,colorlinks]{hyperref}

\usepackage[capitalize]{cleveref}
\crefname{section}{Sec.}{Secs.}
\Crefname{section}{Section}{Sections}
\Crefname{table}{Table}{Tables}
\crefname{table}{Tab.}{Tabs.}

\def\wacvPaperID{1177} 
\def\confName{WACV}
\def\confYear{2025}

\begin{document}

\title{Enhancing Temporal Action Localization:\\Advanced S6 Modeling with Recurrent Mechanism}

\author{Sangyoun Lee\textsuperscript{1}, Juho Jung\textsuperscript{2}, Changdae Oh\textsuperscript{3}, Sunghee Yun\textsuperscript{4}\\
Sogang University\textsuperscript{1}, Sungkyunkwan University\textsuperscript{2}, University of Wisconsin--Madison\textsuperscript{3}, Erudio Bio\textsuperscript{4}\\
{\tt\small leesy0882@sogang.ac.kr\textsuperscript{1}, jhjeon9@g.skku.edu\textsuperscript{2}, changdae.oh@wisc.edu\textsuperscript{3}, sunghee.yun@erudio.bio\textsuperscript{4}}}
\maketitle

\begin{abstract}
   Temporal Action Localization (TAL) is a critical task in video analysis, identifying precise start and end times of actions. Existing methods like CNNs, RNNs, GCNs, and Transformers have limitations in capturing long-range dependencies and temporal causality. To address these challenges, we propose a novel TAL architecture leveraging the Selective State Space Model (S6). Our approach integrates the Feature Aggregated Bi-S6 block, Dual Bi-S6 structure, and a recurrent mechanism to enhance temporal and channel-wise dependency modeling without increasing parameter complexity. Extensive experiments on benchmark datasets demonstrate state-of-the-art results with mAP scores of 74.2\% on THUMOS-14, 42.9\% on ActivityNet, 29.6\% on FineAction, and 45.8\% on HACS. Ablation studies validate our method's effectiveness, showing that the Dual structure in the Stem module and the recurrent mechanism outperform traditional approaches. Our findings demonstrate the potential of S6-based models in TAL tasks, paving the way for future research. Our code is available at \texttt{\href{https://github.com/lsy0882/RDFA-S6}{https://github.com/lsy0882/RDFA-S6}}.

\end{abstract}

\section{Introduction}
\label{sec:intro}

Temporal Action Localization (TAL) is a crucial video analysis task that identifies the precise start and end times of actions in videos. As video content becomes increasingly complex and abundant, accurate TAL methods are essential for effectively capturing and analyzing meaningful actions in applications like sports analytics, surveillance, and interactive media~\cite{idrees2017thumos, caba2015activitynet, liu2022fineaction, zhao2019hacs}. However, significant challenges remain in TAL, particularly in effectively capturing long-range dependencies and temporal causality in video data.

Traditional approaches to TAL, including CNNs, RNNs, GCNs, and Transformers, each bring unique strengths but also have inherent limitations. CNNs are effective at capturing spatial features but struggle with long-range dependencies due to limited receptive fields~\cite{tran2015learning}. RNNs can model temporal sequences but face challenges such as vanishing gradients, which hinder their ability to capture long-term dependencies~\cite{pascanu2013difficulty, gers2000learning}. GCNs are powerful for relational data but are not inherently designed for sequential temporal data~\cite{kipf2016semi}.
Transformers have revolutionized TAL with their ability to model global context using self-attention mechanisms~\cite{vaswani2017attention, Arnab_2021_ICCV}. However, their reliance on attention scores to capture relationships within a sequence does not inherently account for the temporal causality and history of visual elements over time. This limitation makes Transformers less optimal for tasks requiring precise temporal causality, such as TAL, where understanding the sequential nature and dependencies of actions is crucial~\cite{gu2021efficiently, gu2023mamba}.

The State Space Model (SSM)~\cite{gu2021efficiently, gu2023mamba} has emerged as a promising alternative for sequence modeling by addressing the limitations of traditional methods, especially in capturing temporal causality. Within the SSM framework, the Selective State Space Model (S6)~\cite{gu2023mamba} stands out for TAL tasks due to its ability to maintain and leverage historical context through its selection mechanism and gating operation. These properties enable S6 to dynamically adjust the influence of new inputs—specifically, the spatiotemporal feature vectors extracted from the current video clip—ensuring that the model retains and utilizes critical temporal information while integrating new data. This dynamic adjustment and selective retention enable S6 to capture long-range dependencies and temporal causality effectively, providing understanding of action sequences essential for accurately pinpointing the start and end times of actions in TAL.

ActionMamba~\cite{chen2024video}, an S6-based TAL method, has demonstrated that S6-based method can surpass Transformers in sequence modeling by replacing Transformer blocks with S6 blocks. ActionMamba simply substitutes the Transformer-based blocks for sequence modeling in the ActionFormer~\cite{zhang2022actionformer} architecture with S6-based blocks. The S6 blocks use a bi-directional processing approach~\cite{zhu2024vision} and incorporate weight sharing between networks operating in each direction. However, this study lacks a thoughtful design focused on effective TAL methods, instead offering a straightforward replacement of Transformer blocks with slightly enhanced S6 ones. While ActionMamba highlights the potential for S6-based sequence modeling to outperform Transformer-based approaches, it falls short of fully exploring this potential or providing clear guidelines for leveraging S6 effectively in TAL tasks.

Our research aims to explore the potential of S6-based TAL methods by building on insights from previous studies on CNNs, RNNs, GCNs, and Transformers~\cite{tran2015learning, pascanu2013difficulty, gers2000learning, kipf2016semi, vaswani2017attention}. We propose a novel architecture that leverages the strengths of these traditional models while capitalizing on the unique capabilities of S6.

This paper makes the following contributions to the field of TAL:

\begin{enumerate}
    \vspace{-2mm}
    \item \textbf{Advanced Dependency Modeling with S6}: We conduct a pioneering exploration of S6's potential in TAL tasks, particularly focusing on its dependency modeling capabilities. By introducing an advanced dependency modeling technique based on the Feature Aggregated Bi-S6 (FA-Bi-S6) block design and the Dual Bi-S6 structure, we enable robust and effective modeling of dependencies within video sequences. The FA-Bi-S6 block employs multiple Conv1D layers with different kernel sizes to capture various granularities of temporal and channel-wise features, while the Dual Bi-S6 structure processes features along both the temporal and channel dimensions to enhance the integration of spatiotemporal dependencies. This approach provides direction for TAL modeling, enabling more effective utilization of S6 in this domain.
    \vspace{-2mm}
    \item  \textbf{Efficiency through Recurrent Mechanism}: Our study reveals that using a recurrent mechanism to repeatedly apply a single S6-based model outperforms the traditional approach of stacking multiple blocks. This recurrent application enhances the model's performance without increasing the number of parameters, providing an effective solution for improving TAL models.
    \vspace{-2mm}
    \item \textbf{State-of-the-Art Performance}: We achieve state-of-the-art (SOTA) results across multiple benchmark datasets, including THUMOS-14~\cite{idrees2017thumos}, ActivityNet~\cite{caba2015activitynet}, FineAction~\cite{liu2022fineaction}, and HACS~\cite{zhao2019hacs}. Our ablation studies analyze the effectiveness of each component of our proposed architecture, confirming the performance improvements brought by our method.
\end{enumerate}

\section{Related works}
\label{sec:related_works}

\vspace{-2mm}
\paragraph{Convolutional Neural Networks (CNNs)}
Early TAL research used 2D CNNs to process spatial information, with initial attempts like FV-DTF~\cite{oneata2014lear} combining spatial and temporal data but handling them separately. The introduction of 3D CNNs, as seen in CDC~\cite{shou2017cdc}, marked a significant advancement by capturing spatiotemporal features with three-dimensional convolution kernels. However, temporal resolution loss inherent in traditional 3D CNNs was still a challenge to conquer. To cope with this, methods such as TPC~\cite{yang2018exploring} and FSN~\cite{yang2019exploring} aimed to balance spatial and temporal feature processing. GTAN~\cite{long2019gaussian} and PBRNet~\cite{liu2020progressive} further optimized temporal intervals and hierarchical feature extraction. TPC maintained temporal receptive fields while downsampling spatial fields, and FSN captured finer-grained dependencies by sequentially processing spatial and temporal features. 

Our FA-Bi-S6 block builds on these advances by incorporating multiple Conv1D layers with varying kernel sizes in parallel to capture a wide range of local contexts. The resulting feature map is processed bi-directionally by the Bi-S6 network, enhancing the model's ability to capture complex dynamics effectively.

\vspace{-3mm}
\paragraph{Recurrent Neural Networks (RNNs)}
To address the temporal challenges that CNNs alone couldn't solve, RNNs were integrated into TAL frameworks. Early efforts like PSDF~\cite{Yuan_2016_CVPR} and advancements such as AS~\cite{Alwassel_2018_ECCV} used RNNs to enhance temporal context modeling from dense trajectories and refine spatial features for detailed analysis. More sophisticated integrations followed, such as GRU-Split~\cite{keshvarikhojasteh2023temporal}, which employed GRUs to refine action boundaries and probabilities. However, RNNs introduced challenges like managing long sequences and vanishing gradients. RCL~\cite{wang2022rcl} addressed these issues by using a recurrent module to dynamically adjust action segment predictions. 

Our research transcends the limitations of CNNs and RNNs by incorporating a recurrent mechanism within our S6-based architecture. This mechanism, integrated with our Backbone's Stem module, enhances temporal context modeling using the efficiency and precision of state space models.

\vspace{-3mm}
\paragraph{Graph Convolutional Networks (GCNs)}
The limitations of RNNs led to the exploration of GCNs in the TAL domain. GCNs structure video data as graphs, with nodes representing spatiotemporal features and edges defining their relationships, allowing for more comprehensive modeling of temporal dependencies. A notable advancement, P-GCN~\cite{zeng2019graph}, expanded the range of dependencies that could be modeled but faced challenges in scalability and efficiency due to computational overhead. G-TAD~\cite{xu2020g} addressed these issues with a dual-stream graph convolution framework, efficiently capturing both fixed and adaptive temporal dependencies.

Building on GCN insights, we developed the Dual Bi-S6 structure, integrating the TFA-Bi-S6 and CFA-Bi-S6 blocks. TFA-Bi-S6 captures temporal dependencies, while CFA-Bi-S6 handles dependencies between spatiotemporal features by focusing on the channel dimension. This combined approach enhances the robustness and accuracy of TAL by effectively modeling both temporal and channel-wise contexts.

\begin{figure*}
\centering
\begin{subfigure}{\textwidth}
    \centering
    \includegraphics[width=0.97\textwidth]{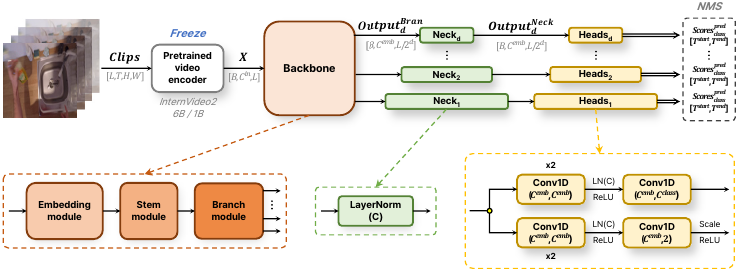}
    \caption{}
    \label{fig:architecture_overview}
\end{subfigure}
\begin{subfigure}{\textwidth}
    \centering
    \includegraphics[width=0.97\textwidth]{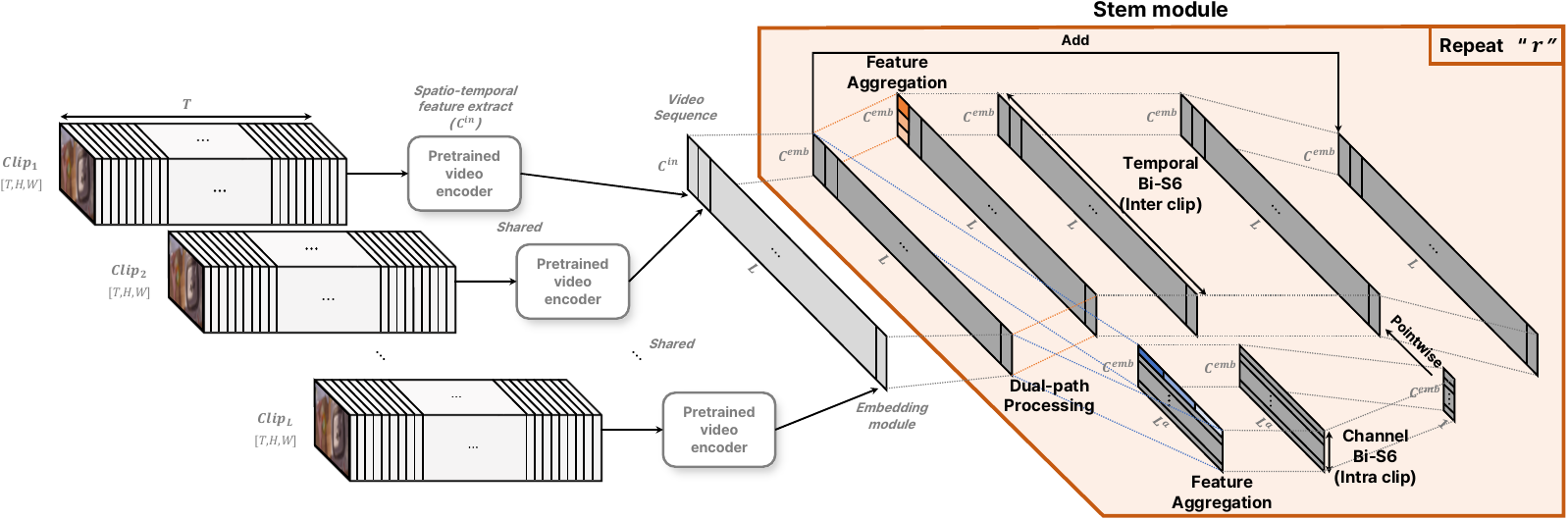}
    \caption{}
    \label{fig:contribution_overview}
\end{subfigure}
\vspace{-5mm}
\caption{\textbf{Illustration of the proposed architecture and its components.} (a) The architecture overview, which consists of four main parts: Pretrained video encoder, Backbone, Neck, and Heads (Action classification head and Temporal boundary regression head). (b) The overview of the proposed methods, highlighting the Stem module with an orange shaded area. The Stem module consists of three parts: Dual-path processing (\textbf{Dual Bi-S6 Structure}), Feature Aggregation \& Temporal/Channel Bi-S6 (\textbf{Feature Aggregated Bi-S6 Block Design}), and the repeat processing with shared networks (\textbf{Recurrent Mechanism}).}
\vspace{-2mm}
\end{figure*}

\vspace{-3mm}
\paragraph{Transformers}
The limitations of GCNs in handling extensive temporal dependencies led to the adoption of Transformer-based models in TAL. Transformers use self-attention to extend temporal dependencies beyond GCN constraints. TRA~\cite{zhao2022temporal} used variable temporal boundary proposals with multi-head self-attention for flexible temporal modeling, though it faced challenges in maintaining temporal causality over long sequences. ActionFormer~\cite{zhang2022actionformer} improved on this by using local self-attention and a multiscale feature pyramid to capture various temporal resolutions, but it still struggled with capturing long-range dependencies and maintaining precise temporal causality.

\vspace{-1mm}
To address these issues, we introduced the S6 network into our TAL system. The S6 network uses selective mechanisms and gating functions to modulate the impact of each time step's spatiotemporal features. This approach allows S6 to preserve critical historical information while integrating new spatiotemporal features, effectively capturing long-range dependencies and temporal causality. By leveraging these capabilities, S6 enhances the accuracy of feature extraction and action localization, addressing the limitations of Transformer-based models in TAL.
\vspace{-1mm}
\section{Proposed Methods}
\label{sec:proposed_methods}

We introduce our approach, emphasizing advanced dependency modeling for TAL by integrating the S6 model to improve long-range dependency handling. Our key components include the \textbf{Feature Aggregated Bi-S6 Block Design}, \textbf{Dual Bi-S6 Structure}, and \textbf{Recurrent Mechanism}.

\subsection{Preliminary: Selective Space State Model (S6)}
Our architecture uses the S6 model with selective mechanisms and gating operations to capture complex temporal dynamics and capture long-range dependencies effectively.

The S6 model operates with parameters \((\Delta_t, A, B, C)\), discretized to manage sequence transformations:
\vspace{-1mm}
\[
h_t = A h_{t-1} + B x_t, \quad y_t = C h_t
\]
Here, \(x_t\) represents the input at time step \(t\), which, in the case of TAL, is the spatiotemporal feature vector extracted from single clip. The hidden state at time step \(t\), \(h_t\), captures the temporal context of the sequence. The output at time step \(t\), \(y_t\), represents the processed feature. The state matrix \(A\) determines how the previous hidden state \(h_{t-1}\) and the historical information from all previous steps influence the current hidden state \(h_t\)~\cite{gu2020hippo}, contributing to precise action localization. The input matrix \(B\) defines how the input \(x_t\) affects the hidden state \(h_t\). Finally, the output matrix \(C\) translates the hidden state \(h_t\) into the output \(y_t\).

The process starts with the input \(x_t\) being projected to derive \(B\), \(C\), and \(\Delta_t\). This step transforms raw input features into suitable representations for state-space modeling. Specifically, the projection functions apply linear transformations to the input \(x_t\):
\vspace{-1mm}
\[B = \text{Linear}(x_t), \quad C = \text{Linear}(x_t)\]

To dynamically manage information flow, the S6 model employs selection mechanism and gating function. The dynamically adjusted parameter \(\Delta_t\) controls the discretization of the state-space model based on the relevance of the input \(x_t\), functioning similarly to a gating mechanism in RNNs. The projection function \(s_\Delta(x_t)\), which includes learnable parameters, projects the input \(x_t\) to one dimension before broadcasting it across channels:
\vspace{-1mm}
\[\Delta_t = \text{softplus}(s_\Delta(x_t))\]

Next, the discretization step adjusts the parameters \(A\) and \(B\) for the current time step \(t\), ensuring that the parameters are appropriately scaled for discrete-time processing:
\vspace{-1mm}
\[
    A_t = \exp(\Delta_t A)
\]
\[
    B_t = (\Delta_t A)^{-1} (\exp(\Delta_t A) - I) \cdot \Delta_t B
\]

The hidden state \(h_t\) is updated using \(A_t\) and \(B_t\), and the output \(y_t\) is generated using \(C_t = C\):
\vspace{-1mm}
\[
    h_t = A_t h_{t-1} + B_t x_t, \quad y_t = C_t h_t
\]

The selective update of the hidden state can be understood as:
\vspace{-1mm}
\[
    h_t = (1 - \Delta_t) h_{t-1} + \Delta_t x_t
\]
where \(\Delta_t\) functions similarly to the gating function \(g_t\) in RNNs, determining the influence of the input \(x_t\) on the hidden state \(h_t\). This dynamic adjustment helps the model focus on relevant portions of the input, ensuring effective handling of long-range dependencies.

S6 is particularly effective in TAL tasks due to its ability to maintain and refine temporal context over extended sequences. By dynamically adjusting \(\Delta_t\), the model can selectively retain important temporal features.

\begin{figure*}
\centering
\subfloat[Embedding module\label{fig:embedding_module}]{\includegraphics[width=0.32\textwidth]{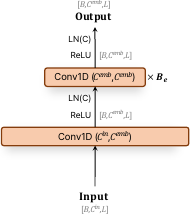}}
\subfloat[Stem module\label{fig:stem_module}]{\includegraphics[width=0.32\textwidth]{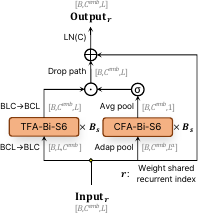}}
\subfloat[Branch module\label{fig:branch_module}]{\includegraphics[width=0.32\textwidth]{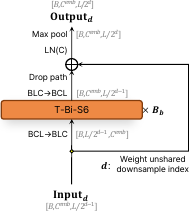}}
\footnotesize
\vspace{-2mm}
\caption{\textbf{Diagrams of the Embedding, Stem, and Branch modules}. (a) Embedding module. (b) Stem module. (c) Branch module.}
\label{fig:backbone_modules}
\end{figure*}
\subsection{Overview}
Our architecture, inspired by ActionFormer~\cite{zhang2022actionformer} and ActionMamba~\cite{chen2024video}, consists of four primary components: a Pretrained video encoder, a Backbone, a Neck, and Heads. The overview of architecture is depicted in Figure \ref{fig:architecture_overview}.

\vspace{-3mm}
\paragraph{Pretrained Video Encoder}
The Pretrained video encoder extracts spatiotemporal attributes from video clips. Trained on diverse datasets such as UCF, Kinetics, Something-Something, and vision-language multi-modal datasets like WebVid and InternVid, it leverages the vast training data from InterVideo2-6B/1B~\cite{wang2024internvideo2}. The pretrained video encoder's example of receiving each clip and extracting spatiotemporal features is shown in Appendix \ref{appendix:example_videoenc}.

\vspace{-3mm}
\paragraph{Backbone}
The Backbone captures dependencies and extracts features at various temporal resolutions from the sequence data. As illustrated in Figure \ref{fig:architecture_overview}, it consists of three main modules:
\begin{itemize}
    \item \textbf{Embedding Module:} This module captures the coarse local context of spatiotemporal features. As shown in Figure \ref{fig:embedding_module}, the sequence is first passed through a Conv1D to increase the dimensionality from $C^{in}$ to $C^{emb}$, followed by Layer Normalization (LN) and ReLU activation. This process is followed by \(B_e\) sequential Conv1D with dimensions $C^{emb}$ to $C^{emb}$, each followed by LN and ReLU activation, resulting in an embedded sequence of shape $[B, C^{emb}, L]$.
    \item \textbf{Stem Module:} This core component processes the embedded sequences to capture long-range dependency using the \textbf{Dual Bi-S6 Structure}. As shown in Figure \ref{fig:stem_module}, it applies two main blocks in parallel: the Temporal Feature Aggregated Bi-S6 (\textbf{TFA-Bi-S6}) block and the Channel Feature Aggregated Bi-S6 (\textbf{CFA-Bi-S6}) block, which focus on capturing temporal and channel-wise dependencies, respectively. Each of these blocks is stacked \(B_s\) times. The TFA-Bi-S6 block handles input sequences reshaped from $[B, C^{emb}, L]$ to $[B, L, C^{emb}]$ and outputs back to $[B, C^{emb}, L]$. The CFA-Bi-S6 block processes the temporal-pooled output of TFA-Bi-S6 with shape $[B, C^{emb}, 1]$ and scales it using a sigmoid activation. The outputs from these blocks are combined through point-wise multiplication with the TFA-Bi-S6 output. This combined output then goes through an affine transformation with a drop path and skip connection, followed by LN to enhance capacity. This process uses a \textbf{Recurrent Mechanism}, repeating \(r\) times, with a weight-shared network applied at each repetition to refine temporal dependency modeling.
    \item \textbf{Branch Module:} This module handles temporal multi-scale dependencies. As shown in Figure \ref{fig:branch_module}, each branch applies the Temporal Bi-S6 (T-Bi-S6) block, which is a modified version of the Bi-S6 block used in ActionMamba \cite{chen2024video}, followed by an affine drop path and residual connection. After this, the output undergoes LN and max pooling along the temporal dimension, effectively obtaining various temporal resolutions. The T-Bi-S6 block processes the input sequence reshaped from $[B, C^{emb}, L/2^{d-1}]$ to $[B, L/2^{d-1}, C^{emb}]$ and outputs back to $[B, C^{emb}, L/2^{d-1}]$. This process is repeated for each downsampling index ($d=1, 2, ..., 5$), where the output shape becomes $[B, C^{emb}, L/2^{d}]$.
\end{itemize}

\vspace{-6mm}
\paragraph{Neck and Heads}
The Neck is designed with simplicity and efficiency in mind, utilizing layer normalization for channel-wise normalization, which is the same as the LN used in the Branch module. This step ensures that the temporal multi-scale sequences reflecting precise temporal dependencies processed by the Backbone are normalized and ready for subsequent processing.

The Heads leverage the normalized features from the Neck to carry out two primary tasks: action classification and temporal boundary regression. The action classification head generates channels equal to the number of action categories, predicting class scores for each category. Simultaneously, the temporal boundary regression head outputs two channels to predict the frame indices marking the start and end of an action. This dual-head design ensures that the model can accurately classify actions and determine their temporal boundaries within the video segments.

\subsection{Advanced Dependency Modeling for TAL}
\begin{figure*}
\centering
\subfloat[TFA-Bi-S6\label{fig:TFA_Bi_S6}]{\includegraphics[width=0.32\textwidth]{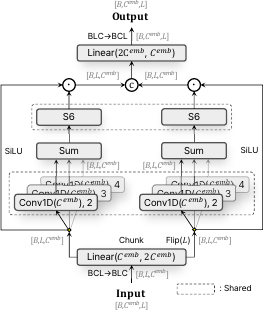}}
\subfloat[CFA-Bi-S6\label{fig:CFA_Bi_S6}]{\includegraphics[width=0.32\textwidth]{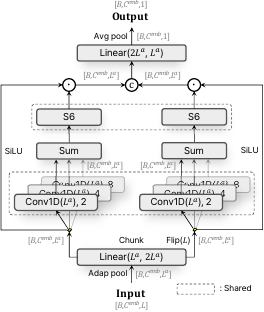}}
\subfloat[T-Bi-S6\label{fig:T_Bi_S6}]{\includegraphics[width=0.32\textwidth]{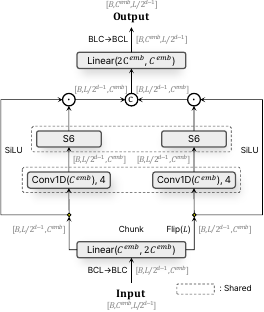}}
\footnotesize
\caption{\textbf{Diagrams of the Feature aggregated Bi-S6 block design}. (a) TFA-Bi-S6 model. (b) CFA-Bi-S6 model. (c) T-Bi-S6 model.}
\label{fig:bi_s6_block_design}
\vspace{-3mm}
\end{figure*}
\paragraph{Feature Aggregated Bi-S6 (FA-Bi-S6) Block Design}
The FA-Bi-S6 block design is one of our contributions, enabling robust and effective modeling of dependencies within video sequences. This block design incorporates multiple Conv1D layers, each with different kernel sizes, operating sequentially within two main blocks: the TFA-Bi-S6 block and the CFA-Bi-S6 block, as shown in Figure \ref{fig:TFA_Bi_S6} and \ref{fig:CFA_Bi_S6}.

In the TFA-Bi-S6 block, the input sequence of shape \([B, C^{emb}, L]\) is first passed through a linear layer that adjusts the dimensions from \([B, C^{emb}, L]\) to \([B, L, 2C^{emb}]\). The sequence is then divided into two chunks, and one of these chunks is flipped. These chunks are processed through multiple Conv1D layers with varying kernel sizes (2, 3, 4), each capturing different granularities of temporal features. The outputs from these Conv1D layers are summed to create an aggregated feature map, which is then processed through a S6 network focusing on temporal dependencies. The output of the S6 blocks is then multiplied pointwise with the original chunked input processed through the SiLU activation. The results from each chunk are concatenated, which handle bi-directional temporal dependencies. The final output is obtained by combining the results, which are then processed through a linear layer and reshaped back to \([B, C^{emb}, L]\).

In the CFA-Bi-S6 block, the process is similar to the TFA-Bi-S6 block with adaptations for channel-wise dependency modeling. The input sequence is first adaptively pooled to \([B, C^{emb}, L^a]\) before the linear layer processing. The Conv1D layers in this block have varying kernel sizes (2, 4, 8) to capture different scales of channel-wise dependencies. After processing through the S6 blocks and linear layer, the final output is average pooled to \([B, C^{emb}, 1]\). These adjustments enable the CFA-Bi-S6 block to focus on capturing diverse channel-wise dependencies and enhance the overall capacity to model complex spatiotemporal interactions within video sequences.

By integrating the Bi-S6 block with the aggregated feature map, our design leverages the strengths of both multi-scale feature extraction and bi-directional processing. The combined architecture allows the model to effectively capture and utilize spatiotemporal features across a wide range of context, addressing the limitations of traditional single convolutional approaches. This design is particularly advantageous for TAL tasks, where actions may occur over varying temporal spans, and the local context provided by surrounding frames is crucial for accurate localization.

\vspace{-3mm}
\paragraph{Dual Bi-S6 Structure}
The Dual Bi-S6 structure is a novel component of our proposed architecture, designed to enhance the modeling of spatiotemporal dependencies by processing features along both the temporal and channel dimensions. This dual-path approach ensures that the model can capture and integrate the rich contextual information present in video sequences, thereby improving the accuracy of TAL.

As shown in Figure \ref{fig:stem_module}, the Dual Bi-S6 structure consists of two parallel paths: the TFA-Bi-S6 and the CFA-Bi-S6. Each path processes the input sequence differently to extract complementary information.
The TFA-Bi-S6 reflects temporal dynamics within the video sequence, providing a detailed temporal analysis of the input.
Simultaneously, the CFA-Bi-S6 captures the interactions between different spatiotemporal features, and its output is then scaled using a sigmoid function to transform the values into a range suitable for modulation.

After processing the input through both paths, the outputs of the TFA-Bi-S6 and CFA-Bi-S6 are combined using point-wise multiplication. This fusion step integrates the temporal dependencies captured by the TFA-Bi-S6 with the channel-wise dependencies modeled by the CFA-Bi-S6. The point-wise multiplication ensures that the combined features reflect both types of dependencies, with the TFA-Bi-S6 handling global dependencies between clips and the CFA-Bi-S6 addressing local dependencies between spatiotemporal features within clips. The design intention behind this structure is to leverage the strengths of both paths: the TFA-Bi-S6 captures temporal dependencies and dynamics, while the CFA-Bi-S6 emphasizes the relationships between spatiotemporal features. By scaling the output of the CFA-Bi-S6 and multiplying it with the TFA-Bi-S6 output, the model effectively combines temporal analysis with channel-wise context, leading to a more comprehensive understanding of the video.

\vspace{-3mm}
\paragraph{Recurrent Mechanism}
This mechanism, integrated with our Stem module in the Backbone, enhances the accuracy of temporal context modeling by leveraging the efficiency and precision of state space models. As shown in Figure \ref{fig:stem_module}, the process begins by passing the input sequence through the Stem module to capture initial temporal dependencies. The output is combined with the original input sequence and reprocessed by the Stem module, repeating this process \(r\) times. Each iteration refines the temporal dependencies further, enhancing the model's ability to capture long-range dependencies and intricate temporal patterns. This recurrent mechanism provides a robust framework for refining temporal context, allowing the model to improve its understanding of temporal dependencies dynamically.

The effectiveness of this recurrent mechanism in speech separation tasks highlights its potential for TAL tasks as well. In speech separation, recurrent mechanisms have proven to excel in capturing long-range dependencies and intricate temporal patterns~\cite{hu2021speech, chen2023neural}. This iterative refinement process, which involves passing the input sequence through a module multiple times to capture and refine temporal dependencies, allows models to handle complex long-range dependencies with greater precision. Such capabilities are directly applicable to TAL tasks, where identifying precise segments within a video also requires understanding temporal dependencies over extended periods. 
\label{sec:total_eval}
\begin{table*}[!htb]
\centering
\footnotesize
\renewcommand{\tabcolsep}{3.5pt}
\def\arraystretch{0.98}
\setlength{\extrarowheight}{0pt}

\begin{subtable}[t]{0.48\textwidth}
\centering
\begin{tabular}{lccccccc}
\multirow{2}[2]{*}{\textbf{Base}} & \multirow{2}[2]{*}{\textbf{System}} &  \multicolumn{6}{c}{\textbf{mAP} (\%)} \\
\cmidrule(lr){3-8}
 & & \textbf{@.3} & \textbf{@.4} & \textbf{@.5} & \textbf{@.6} & \textbf{@.7} & \textbf{Avg} \\
\Xhline{2.5\arrayrulewidth}
CNN & CDC~\cite{shou2017cdc} & 40.1 & 29.4 & 23.3 & 13.1 & 7.9 & 22.8 \\
 & TAL-Net~\cite{chao2018rethinking} & 53.2 & 48.5 & 42.8 & 33.8 & 20.8 & 39.8 \\
 & PBRNet~\cite{liu2020progressive} & 58.5 & 54.6 & 51.3 & 41.8 & 29.5 & 47.1 \\
\Xhline{2.5\arrayrulewidth}
RNN & AS~\cite{Alwassel_2018_ECCV} & 51.8 & 42.4 & 30.8 & 20.2 & 11.1 & 31.3 \\
 & RCL~\cite{wang2022rcl} & 70.1 & 62.3 & 52.9 & 42.7 & 30.7 & 51.7 \\
\Xhline{2.5\arrayrulewidth}
GCN & G-TAD~\cite{xu2020g} & 66.4 & 60.4 & 51.6 & 37.6 & 22.9 & 47.8 \\
\Xhline{2.5\arrayrulewidth}
Transformer & TallFormer~\cite{cheng2022tallformer} & 76.0 & 71.5 & 63.2 & 50.9 & 34.5 & 59.2 \\
 & ActionFormer~\cite{zhang2022actionformer} & 82.1 & 77.8 & 71.0 & 59.4 & 43.9 & 66.8 \\
 & TriDet~\cite{shi2023tridet} & 83.6 & 80.1 & 72.9 & 62.4 & 47.4 & 69.3 \\
\Xhline{2.5\arrayrulewidth}
S6 & ActionMamba~\cite{chen2024video} & 86.9 & 83.1 & 76.9 & 65.1 & 50.8 & 72.7 \\
\rowcolor{Gray}
 & Ours & 88.7 & 84.6 & 78.2 & 66.6 & 51.9 & 74.2 \\
\Xhline{2.5\arrayrulewidth}
\end{tabular}
\caption{}
\label{tab:benchmark_thumos}
\end{subtable}
\begin{subtable}[t]{0.48\textwidth}
\centering
\begin{tabular}{lccccc}
\multirow{2}[2]{*}{\textbf{Base}} & \multirow{2}[2]{*}{\textbf{System}} &  \multicolumn{4}{c}{\textbf{mAP} (\%)} \\
\cmidrule(lr){3-6}
 & & \textbf{@.5} & \textbf{@.75} & \textbf{@.95} & \textbf{Avg} \\
\Xhline{2.5\arrayrulewidth}
CNN & BSN~\cite{lin2018bsn} & 46.5 & 30.0 & 8.0 & 30.0 \\
 & DCAN~\cite{chen2022dcan} & 51.8 & 36.0 & 9.5 & 35.4 \\
\Xhline{2.5\arrayrulewidth}
RNN & DeepAct~\cite{song2018deepact} & 37.8 & 24.8 & 10.0 & 24.0 \\
\Xhline{2.5\arrayrulewidth}
GCN & G-TAD~\cite{xu2020g} & 50.4 & 34.6 & 9.0 & 34.1 \\
 & AVFusion~\cite{bagchi2021hear} & 54.3 & 37.7 & 8.9 & 36.8 \\
\Xhline{2.5\arrayrulewidth}
Transformer & ActionFormer~\cite{zhang2022actionformer} & 54.7 & 37.8 & 8.4 & 36.6 \\
 & TriDet~\cite{shi2023tridet} & 54.7 & 38.0 & 8.4 & 36.8 \\
 & TCANet~\cite{qing2021temporal} & 54.3 & 39.1 & 8.4 & 37.6 \\
 & AdaTAD~\cite{liu2024end} & 61.7 & 43.4 & 10.9 & 41.9 \\
\Xhline{2.5\arrayrulewidth}
S6 & ActionMamba~\cite{chen2024video} & 62.4 & 43.5 & 10.2 & 42.0 \\
\rowcolor{Gray}
 & Ours & 64.1 & 44.0 & 10.6 & 42.9 \\
\Xhline{2.5\arrayrulewidth}
\end{tabular}
\caption{}
\label{tab:benchmark_activitynet}
\end{subtable}
\begin{subtable}[t]{0.48\textwidth}
\centering
\begin{tabular}{lccccc}
\multirow{2}[2]{*}{\textbf{Base}} & \multirow{2}[2]{*}{\textbf{System}} &  \multicolumn{4}{c}{\textbf{mAP} (\%)} \\
\cmidrule(lr){3-6}
 & & \textbf{@.5} & \textbf{@.75} & \textbf{@.95} & \textbf{Avg} \\
\Xhline{2.5\arrayrulewidth}
CNN & DBG~\cite{lin2020fast} & 10.7 & 6.4 & 2.5 & 6.8 \\
\Xhline{2.5\arrayrulewidth}
GCN & G-TAD~\cite{xu2020g} & 13.7 & 8.8 & 3.1 & 9.1 \\
\Xhline{2.5\arrayrulewidth}
Transformer & VideoMAE-v2~\cite{wang2023videomae} & 29.1 & 17.7 & 5.1 & 18.2 \\
\Xhline{2.5\arrayrulewidth}
S6 & ActionMamba~\cite{chen2024video} & 45.4 & 28.8 & 6.8 & 29.0 \\
\rowcolor{Gray}
 & Ours & 46.4 & 29.5 & 7.6 & 29.6 \\
\Xhline{2.5\arrayrulewidth}
\end{tabular}
\caption{}
\label{tab:benchmark_fineaction}
\end{subtable}
\begin{subtable}[t]{0.48\textwidth}
\centering
\begin{tabular}{lccccc}
\multirow{2}[2]{*}{\textbf{Base}} & \multirow{2}[2]{*}{\textbf{System}} &  \multicolumn{4}{c}{\textbf{mAP} (\%)} \\
\cmidrule(lr){3-6}
 & & \textbf{@.5} & \textbf{@.75} & \textbf{@.95} & \textbf{Avg} \\
\Xhline{2.5\arrayrulewidth}
CNN & DyFADet~\cite{yang2024dyfadet} & 64.0 & 44.8 & 14.1 & 44.3 \\
\Xhline{2.5\arrayrulewidth}
Transformer & TadTR~\cite{liu2022end} & 47.1 & 32.1 & 10.9 & 32.1 \\
 & TriDet~\cite{shi2023tridet} & 62.4 & 44.1 & 13.1 & 43.1 \\
\Xhline{2.5\arrayrulewidth}
S6 & ActionMamba~\cite{chen2024video} & 64.0 & 45.7 & 13.3 & 44.6 \\
\rowcolor{Gray}
 & Ours & 66.4 & 47.2 & 14.3 & 45.8 \\
\Xhline{2.5\arrayrulewidth}
\end{tabular}
\caption{}
\label{tab:benchmark_hacs}
\end{subtable}
\vspace{-2mm}
\caption{\textbf{Results of temporal action localization on benchmark datasets.} (a) THUMOS-14~\cite{idrees2017thumos}, (b) ActivityNet~\cite{caba2015activitynet}, (c) FineAction~\cite{liu2022fineaction}, (d) HACS~\cite{zhao2019hacs}. The metric used is mean Average Precision (mAP) evaluated at multiple tIoU thresholds.}
\label{tab:benchmark_results}
\end{table*}

\section{Experiments}
\label{sec:experiments}
\vspace{-2mm}
We provide a comprehensive evaluation of our TAL method through extensive experiments. We demonstrate its effectiveness using various benchmark datasets and conduct ablation studies to assess the impact of various components of our proposed approach.

\vspace{-1mm}
\subsection{Evaluation on Benchmarks}

To evaluate the effectiveness of the proposed method for TAL, we utilized the benchmark datasets THUMOS-14~\cite{idrees2017thumos}, ActivityNet~\cite{caba2015activitynet}, FineAction~\cite{liu2022fineaction}, and HACS~\cite{zhao2019hacs}. Detailed descriptions of each benchmark can be found in Appendix~\ref{appendix:benchmarks_info}.

Table~\ref{tab:benchmark_thumos} presents experimental results on THUMOS-14. We compared our method with various approaches, including CNNs, RNNs, GCNs, Transformers-based, and the latest SOTA S6-based model. Our method achieved an average mAP of 74.2\%, surpassing the previous SOTA by 1.5\%. In Table~\ref{tab:benchmark_activitynet}, we summarize our performance on ActivityNet. Despite its larger scale and variety of classes, which generally result in lower scores, our method achieved an average mAP of 42.9\%, surpassing the previous SOTA by 0.9\%.

The outcomes on FineAction are presented in Table~\ref{tab:benchmark_fineaction}. This benchmark, being relatively new, lacked RNN-based studies for comparison. Therefore, we included studies utilizing CNN, GCN, Transformer, and S6 models. FineAction's high class variety relative to its size makes it particularly challenging, generally resulting in lower mAP scores. Nonetheless, our approach achieved an average mAP of 29.6\%, which is 0.6\% higher than the previous SOTA. Finally, Table~\ref{tab:benchmark_hacs} displays our experimental performance on HACS. Most studies focused on Transformer-based approaches due to the dataset's large scale. Despite this, our proposed method achieved an average mAP of 45.8\%, exceeding the previous SOTA by 1.2\%.

\label{sec:ablation_studies}
\begin{table*}[!htb]
\centering
\begin{minipage}{0.32\textwidth}
\centering
\footnotesize
\renewcommand{\tabcolsep}{1.5pt}
\def\arraystretch{1.06}
\begin{subtable}{\textwidth}
\centering
\begin{tabular}{lccc}
\multirow{2}{*}{\textbf{Structure}} & \multirow{2}{*}{\textbf{($B_e$,$B_s$,$B_b$)}} & \multirow{1}{*}{\textbf{Params}} & \multirow{1}{*}{\textbf{Avg mAP}} \\
 &  & \textbf{(M)} & \textbf{(\%)} \\
\Xhline{2.5\arrayrulewidth}
Single & {(1,0,1)} & {16.0} & {69.4} \\
 & {(1,1,1)} & {18.8} & {72.2} \\
 & {(2,1,1)} & {19.6} & {72.0} \\
 & {(1,2,1)} & {21.6} & {71.7} \\
 & {(1,1,2)} & {33.0} & {71.0} \\
 & {(2,2,1)} & {22.5} & {71.8} \\
 & {(2,1,2)} & {33.8} & {71.1} \\
 & {(1,2,2)} & {35.9} & {71.3} \\
 & {(2,2,2)} & {36.6} & {71.5} \\
 & {(1,4,1)} & {27.3} & {71.3} \\
 & {(1,8,1)} & {38.6} & {70.7} \\
\Xhline{2.5\arrayrulewidth}
\rowcolor{Gray}
Dual & {(1,1,1)} & {21.7} & {72.8} \\
 & {(1,2,1)} & {28.5} & {72.5} \\
\Xhline{2.5\arrayrulewidth}
\end{tabular}
\caption{}
\label{tab:ablation_structure}
\end{subtable}
\end{minipage}
\begin{minipage}{0.32\textwidth}
\centering
\footnotesize
\renewcommand{\tabcolsep}{1.5pt}
\def\arraystretch{1.06}
\begin{subtable}{\textwidth}
\centering
\begin{tabular}{ccccc}
\multirow{2}{*}{\textbf{$K_{TFA}$}} & \multirow{2}{*}{\textbf{$K_{CFA}$}} & \multirow{2}{*}{\textbf{Aggregate}} & \multirow{1}{*}{\textbf{Params}} & \multirow{1}{*}{\textbf{Avg mAP}} \\
 & & & \textbf{(M)} & \textbf{(\%)} \\
\Xhline{2.5\arrayrulewidth}
{X} & {X} & {Sum} & {20.5} & {72.1} \\
{(4)} & {X} & {Sum} & {21.6} & {72.6} \\
{X} & {(4)} & {Sum} & {20.6} & {72.3} \\
{(4)} & {(4)} & {Sum} & {21.7} & {72.8} \\
{(2,4)} & {(4)} & {Sum} & {22.1} & {73.0} \\
{(4)} & {(2,4)} & {Sum} & {21.7} & {72.9} \\
{(2,4)} & {(2,4)} & {Sum} & {22.2} & {73.1} \\
{(2,3,4)} & {(2,3,4)} & {Sum} & {23.0} & {73.4} \\
{(2,3,4,8)} & {(2,3,4,8)} & {Sum} & {25.2} & {73.2} \\
\Xhline{2.5\arrayrulewidth}
{(2,4,8)} & {(2,3,4)} & {Sum} & {24.3} & {73.4} \\
\rowcolor{Gray}
{(2,3,4)} & {(2,4,8)} & {Sum} & {23.1} & {73.5} \\
{(2,3,4)} & {(2,4,8)} & {Concat} & {31.6} & {72.8} \\
{(2,4,8)} & {(2,4,8)} & {Sum} & {24.5} & {73.4} \\
\Xhline{2.5\arrayrulewidth}
\end{tabular}
\caption{}
\label{tab:ablation_kernels}
\end{subtable}
\end{minipage}
\begin{minipage}{0.32\textwidth}
\centering
\footnotesize
\renewcommand{\tabcolsep}{7.5pt}
\def\arraystretch{1.06}
\begin{subtable}{\textwidth}
\centering
\begin{tabular}{cccc}
\multirow{2}{*}{\textbf{$B_s$}} & \multirow{2}{*}{\textbf{$r$}} & \multirow{1}{*}{\textbf{Params}} & \multirow{1}{*}{\textbf{Avg mAP}} \\
 &  & \textbf{(M)} & \textbf{(\%)} \\
\Xhline{2.5\arrayrulewidth}
1 & {1} & {23.1} & {73.5} \\
 & {2} & {23.1} & {73.6} \\
 & {4} & {23.1} & {73.7} \\
 & {8} & {23.1} & {73.9} \\
\rowcolor{Gray}
 & {16} & {23.1} & {74.2} \\
 & {32} & {23.1} & {74.0} \\
\Xhline{2.5\arrayrulewidth}
2 & {1} & {31.3} & {73.1} \\
 & {16} & {31.3} & {73.4} \\
 & {32} & {31.3} & {73.2} \\
\Xhline{2.5\arrayrulewidth}
4 & {1} & {47.8} & {72.8} \\
 & {16} & {47.8} & {72.6} \\
 & {32} & {47.8} & {72.1} \\
\Xhline{2.5\arrayrulewidth}
8 & {1} & {80.9} & {72.3} \\
\Xhline{2.5\arrayrulewidth}
\end{tabular}
\caption{}
\label{tab:ablation_recurrent}
\end{subtable}
\end{minipage}
\vspace{-3mm}
\caption{\textbf{Ablation studies on the proposed methods.} (a) Performance comparison with varying numbers of blocks in the Embedding, Stem, and Branch modules ($B_e$, $B_s$, $B_b$) and different structures (Structure) using only single Conv1D layer without Feature Aggregation. In this context, ``Single'' refers to using only the temporal block in the Stem module, while ``Dual'' refers to using both the temporal and channel blocks in the Stem module. (b) Performance comparison with different kernel size combinations for TFA-Bi-S6 and CFA-Bi-S6 blocks ($K_{TFA}$ and $K_{CFA}$) and different aggregation methods (Aggregate) using the Dual structure. (c) Performance comparison with varying iterations ($r$) of applying residual connections in the recurrent Dual S6 structure in the Stem module and different numbers of blocks in the Stem module ($B_s$), with both Dual structure and Feature Aggregation applied. All results are from the THUMOS-14 dataset.}
\label{tab:ablation_studies}
\end{table*}

\vspace{-1mm}
\subsection{Ablation Studies}
\paragraph{Stem module structure and Block quantities}
We investigated the impact of varying the \emph{\textbf{structure of the Stem module}} and the \emph{\textbf{number of blocks}} in the Embedding, Stem, and Branch modules to understand their effect on performance.

The results, presented in Table~\ref{tab:ablation_structure}, demonstrate the superiority of the Dual structure in the Stem module, which utilizes both temporal and channel blocks, consistently outperforming the Single structure that only uses the temporal block. This finding suggests that addressing both temporal and channel-wise dependencies provides a more comprehensive understanding for TAL. Additionally, using a single block in each module often yielded better performance than multiple blocks, indicating that simpler, less complex model structures help prevent overfitting and effectively capture essential spatiotemporal features. Notably, omitting the Stem module ($B_s=0$) results in a significant performance drop, highlighting its importance in sequence interpretation.

\vspace{-3mm}
\paragraph{Kernel sizes and Aggregation methods}
We evaluated the performance impact of different \emph{\textbf{kernel size}} combinations for TFA-Bi-S6 and CFA-Bi-S6 blocks and various \emph{\textbf{aggregation methods}} using the Dual structure. This analysis, detailed in Table~\ref{tab:ablation_kernels}, explores how different configurations influence the model's ability to capture temporal and channel-wise local context.

The results show that using multiple kernel sizes for Conv1D layers in both TFA-Bi-S6 and CFA-Bi-S6 blocks improves performance, demonstrating the benefit of capturing a diverse range of local contexts at multiple scales for TAL. However, configurations with four or more kernel sizes per block resulted in decreased performance, likely due to overfitting, as the increased model complexity led to learning noise and less relevant patterns.

The absence of Conv1D layers led to reduced performance, underscoring the importance of capturing temporal and channel-wise local context through these layers. Furthermore, the Sum aggregation method outperformed the Concat method, indicating that summing feature maps effectively integrates information across different scales without adding excessive complexity.

\vspace{-2mm}
\paragraph{Recurrent mechanism iterations}
We examined the impact of varying the number of iterations $r$ in the \emph{\textbf{recurrent mechanism}}, along with the \emph{Dual structure} and \emph{Feature Aggregation}. This analysis, detailed in Table~\ref{tab:ablation_recurrent}, assesses how iterative refinement of temporal dependencies affects model performance compared to increasing the number of Stem blocks ($B_s$).

The results show that increasing the number of recurrent iterations $r$ generally improves performance up to a certain point. Beyond this point, however, additional iterations resulted in a slight performance drop, likely due to an imbalance in temporal dependency. This suggests that there is an optimal number of iterations after which the benefits begin to diminish. In contrast, increasing the number of Stem blocks ($B_s$) while keeping $r$ fixed at 1 led to a decrease in performance, indicating that simply adding more Stem blocks is not effective for improving TAL.

This comparison shows that adopting a recurrent approach, with $B_s$ set to 1 and increasing $r$, is more efficient and effective than stacking additional blocks. The recurrent mechanism improves temporal precision and long-range dependency modeling while optimizing memory usage, crucial for accurately understanding extended actions in video sequences and boosting performance, making it a practical strategy for TAL tasks using the S6-based model.
\vspace{-1mm}
\section{Conclusion}
\label{sec:conclusion}
In this paper, we introduced a novel architecture leveraging S6 to provide effective solutions for TAL tasks based on insights from previous studies. By integrating the Feature Aggregated Bi-S6 block and the Dual Bi-S6 structure, our approach captures multi-scale temporal and channel-wise dependencies. The recurrent mechanism further refines temporal context modeling, enhancing performance without increasing parameter complexity. Consequently, our approach achieves state-of-the-art results on various benchmark datasets, with average mAP scores of 74.2\% on THUMOS-14, 42.9\% on ActivityNet, 29.6\% on FineAction, and 45.8\% on HACS. Additionally, ablation studies confirm the advantages of our design, demonstrating that the Dual structure in the Stem module outperforms the Single structure, the recurrent mechanism is more effective than merely stacking additional blocks, and Temporal Aggregation further boosts performance.  These findings pave the way for future research to further explore the potential of state space models in TAL tasks.

{\small
\bibliographystyle{ieee_fullname}
\bibliography{egbib}
}
\vspace{110mm}

\begin{figure*}[ht]
\centering
\includegraphics[width=0.97\textwidth]{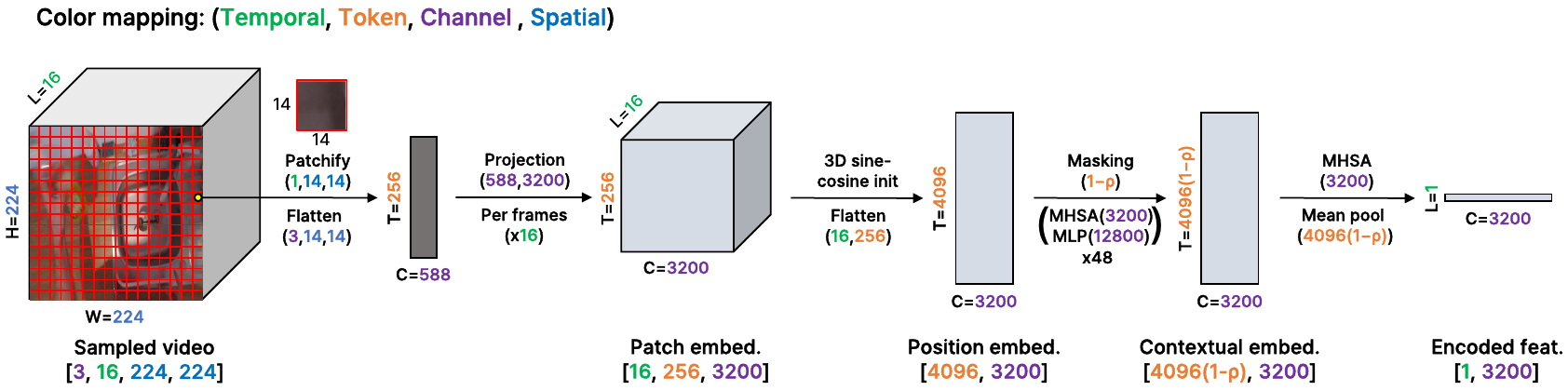}
\caption{\textbf{Process of extracting spatiotemporal features using the pretrained video encoder.} The encoder processes 30fps RGB video frames, groups them into 16-frame clips, and applies patchification, positional embedding, and multi-head self-attention to produce encoded feature vectors.}
\label{fig:pretrain_encoder}
\vspace{-3mm}
\end{figure*}
\appendix

\section*{Appendix / supplemental material}
This appendix is organized as follows:
\begin{itemize}
\small
\item \textbf{Appendix A} provides a detailed explanation of the process involved in extracting spatiotemporal features from video clips using a pretrained video encoder. It covers the technical steps and methodology used, including patchification, positional embedding, and multi-head self-attention mechanisms.
\item \textbf{Appendix B} describes the benchmark datasets used for evaluating Temporal Action Localization (TAL) methods. It includes detailed descriptions of datasets such as THUMOS-14, ActivityNet, FineAction, and HACS, along with their key characteristics and evaluation metrics.
\end{itemize}

\section{Example of Pretrained Video Encoder Extracting Spatiotemporal Features from Each Clip}
\label{appendix:example_videoenc}

To understand the design intention of the Dual Bi-S6 structure, it is crucial to explain how the Pretrained video encoder extracts spatiotemporal features from each clip, clarifying the information contained in the sequence. 

For instance, when processing the THUMOS dataset using the same Pretrained video encoder as ActionMamba, we start with RGB videos at 30 fps and a spatial resolution of 224x224. We segment 16 frames into a single clip, setting a frame interval of 4 (stride=4) between clips, yielding multiple clips from each video, each clip measuring $[3, 16, 224, 224]$. Within each frame, patches of size 14x14 are generated, producing 256 patch tokens per frame. Each patch token, representing spatial information and RGB channels, is flattened to a dimension of $[256, 588]$. These spatial tokens are projected to a channel size of 3200, forming patch embedding tokens with dimensions $[16, 256, 3200]$. Adding 3D sine-cosine positional embeddings to both the patch and frame dimensions, and then flattening these dimensions, results in position-embedded tokens with dimensions $[4096, 3200]$. Next, a proportion $\rho$ of tokens is masked, and the channels are projected to 3200, followed by multi-head self-attention and a feedforward layer with a hidden channel size of 12800, repeated 48 times to incorporate spatiotemporal context, resulting in contextual embedded tokens with dimensions $[4096(1-\rho), 3200]$. Finally, multi-head self-attention and mean pooling are applied along the token dimension to produce an encoded feature vector with dimensions $[1, 3200]$ for each clip. This process is repeated for all clips, stacking the encoded feature vectors sequentially over time to generate the sequence data, excluding the first and last two clips, which may lack video information, as shown in Figure \ref{fig:pretrain_encoder}.

\section{Benchmark Datasets for Temporal Action Localization}
\label{appendix:benchmarks_info}

To provide a comprehensive evaluation of TAL methods, we employ several benchmark datasets that vary in size, complexity, and focus. Here, we describe the key characteristics and evaluation metrics of the datasets utilized in this study:

\textbf{THUMOS-14}: This large-scale dataset is specifically designed for video action recognition and includes detailed temporal frame index annotations for 20 action classes. The primary evaluation metric for THUMOS-14 is mean Average Precision (mAP), which is calculated at various temporal Intersection over Union (tIoU) thresholds of 0.3, 0.4, 0.5, 0.6, and 0.7. This allows for a thorough assessment of the model's performance across different levels of temporal precision.

\textbf{ActivityNet}: Significantly larger and more complex than THUMOS-14, ActivityNet comprises approximately 20,000 videos spanning 200 action classes. The diverse range of classes in ActivityNet presents a more challenging scenario for TAL models. The mAP evaluation metric is also employed here, with tIoU thresholds set at 0.5, 0.75, and 0.95, providing a stringent test for action localization performance.

\textbf{FineAction}: Consisting of around 16,000 videos featuring 106 action classes, FineAction emphasizes everyday activities and sports. The high variety of classes relative to its size makes it a particularly challenging dataset. Evaluation methods are akin to those used for ActivityNet, utilizing mAP scores at multiple tIoU thresholds.

\textbf{HACS (Human Action Clips and Segments)}: This extensive dataset includes approximately 50,000 videos covering 200 action classes, primarily capturing various actions from everyday life. Evaluation of the HACS dataset is conducted using the same methodology as ActivityNet, ensuring a consistent benchmark for comparing TAL model performance across different datasets.

These detailed descriptions of the datasets underscore the diverse and comprehensive nature of the benchmarks used in this study, providing a robust framework for evaluating the effectiveness of TAL methods.
\end{document}